\documentclass[11pt,a4paper]{article}
\usepackage[hyperref]{acl2018}
\usepackage{times}
\usepackage{latexsym}

\usepackage{url}

\usepackage{amsfonts}
\usepackage{booktabs}
\usepackage{graphicx}

\aclfinalcopy 
\def\aclpaperid{1470} 

\usepackage{array}
\newcolumntype{H}{>{\setbox0=\hbox\bgroup}c<{\egroup}@{}}

\newcommand{\Ni}{({\em i})~}
\newcommand{\Nii}{({\em ii})~}
\newcommand{\Niii}{({\em iii})~}
\newcommand{\Niv}{({\em iv})~}

\title{Injecting Relational Structural Representation in Neural Networks\\
 for Question Similarity}
 
\author{ Antonio Uva$^{\dagger}$ \and Daniele Bonadiman$^{\dagger}$ \and Alessandro Moschitti  \\
 $^{\dagger}$DISI, University of Trento, 38123 Povo (TN), Italy \\
 Amazon, Manhattan Beach, CA, USA, 90266 \\ 
\url{{antonio.uva, d.bonadiman}@unitn.it}\\ \url{amosch@amazon.com}
}

\date{}

\begin{document}
\maketitle
\begin{abstract}

Effectively using full syntactic parsing information in Neural Networks (NNs) to solve relational tasks, e.g., question similarity, is still an open problem. In this paper, we propose to inject structural representations in NNs by \Ni learning an SVM model using Tree Kernels (TKs) on relatively few pairs of questions (few thousands) as gold standard (GS) training data is typically scarce, \Nii predicting labels on a very large corpus of question pairs, and \Niii pre-training NNs on such large corpus. The results on Quora and SemEval question similarity datasets show that NNs trained with our approach can learn more accurate models, especially after fine tuning on  GS.


\end{abstract}

\section{Introduction}

Recent years have seen an exponential growth and use of web forums, where users can exchange and find information just asking questions in natural language. Clearly, the possibility of reusing previously asked questions makes  forums much more useful. Thus, many tasks have been proposed to build automatic systems for detecting duplicate questions. These were both organized in academia, e.g., SemEval \cite{S16-1083,S17-2003}, or companies, e.g., Quora  \footnote{https://www.kaggle.com/c/quora-question-pairs}. 
An interesting outcome of the SemEval challenge was that syntactic information is essential to achieve high accuracy in  question reranking tasks. Indeed, the top-systems were built using Support Vector Machines (SVMs) trained with Tree Kernels (TKs), which were applied to a syntactic representation of question text \cite{S16-1172, S17-2053,S16-1138}.

In contrast, NNs-based models struggled to get good accuracy as \Ni large training sets are typically not available \footnote{SQuAD by \citet{D16-1264} is an exception, also possible because dealing with a simpler factoid QA task}, and \Nii effectively exploiting full-syntactic parse information in NNs is still an open issue. 
Indeed, despite \citet{das-EtAl:2016:P16-1} showed that NNs are very effective to manage lexical variability, no neural model encoding syntactic information has shown a clear improvement. Indeed, also NNs directly exploiting syntactic information, such as the Recursive Neural Networks by \citet{socher2013recursive} or the Tree-LSTM by \citet{tai2015improved}, have been shown to be outperformed by well-trained sequential models \cite{li2015tree}. 

Finally, such tree-based approaches depend on sentence structure, thus are difficult to optimize and parallelize.
%
%
%
%
This is a shame as NNs are very flexible in general and enable an easy system deployment in real applications, while TK models require syntactic parsing and longer testing time.
%

In this paper, we propose an approach that aims at injecting syntactic information in NNs, still keeping them simple. It consists of the following steps: \Ni train a TK-based model on a few thousands training examples; 
\Nii apply such classifier to a much larger set of unlabeled training examples to generate automatic annotation; \Niii pre-train NNs on the automatic data; and \Niv fine-tune NNs on the smaller 
GS data. 

Our experiments on two different datasets, i.e., Quora and Qatar Living (QL) from SemEval, show that \Ni when NNs are pre-trained on the predicted data, they achieve accuracy higher than the one of TK models and \Nii NNs can be further boosted by fine-tuning them on the available GS data. This suggests that the TK properties are captured by NNs, which can exploit syntactic information even more effectively, thanks to their well-known generalization ability.

In contrast to other semi-supervised approaches, e.g., self-training, we show that the improvement of our approach is obtained only when a very different classifier, i.e., TK-based, is used to label a large portion of the data. Indeed, using the same NNs in a self-training fashion (or another NN in a co-training approach) to label the semi-supervised data does not provide any improvement. Similarly, when SVMs using standard similarity lexical features are applied to label data, no improvement is observed in NNs. 

One evident consideration is the fact that TKs-based models mainly exploit syntactic information to classify data. Although, assessing that NNs specifically learn such syntax should require further investigation, our results show that only the transfer from TKs produces improvement: this is a significant evidence that makes it worth to further investigate the main claim of our paper. In any case, our approach increases the accuracy of NNs, when small datasets are available to learn high-level semantic task such as question similarity. It consists in \Ni using heavier syntactic/semantic models, e.g., based on TKs, to produce training data; and \Nii exploit the latter to learn a neural model, which can then be fine-tuned on the small available GS data.


\section{Tasks and Baseline Models}

We introduce our question similarity tasks along with two of the most competitive models for their solutions.

\subsection{Question Matching and Ranking}

Question similarity in forums can be set in different ways, e.g., detecting if two questions are semantically similar or ranking a set of retrieved questions in terms of their similarity with the original question. We describe the two methods below:

The Quora task regards detecting if two questions are duplicate or not, or, in other words, if they have the same intent. The associated dataset \cite{wang2017bilateral} contains over $404,348$ pairs of questions, posted by users on the Quora website, labelled as duplicate pair or not. For example, \textit{How do you start a bakery?} and \textit{How can one start a bakery business?}  are duplicated while \textit{What are natural numbers?} and \textit{What is a least natural number?} are not. The ground-truth labels contain some amount of noise.

In the QL task at SemEval-2016 \cite{S16-1083} users were provided with a new (original) question $q_o$ and a set of related questions  $(q_1, q_2, ... q_n)$ from the QL forum\footnote{http://www.qatarliving.com/forum} retrieved by a search engine, i.e., Google. The goal is to rank question candidates, $q_i$, by their similarity with respect to $q_o$. 
$q_i$ were manually annotated as \emph{PerfectMatch}, \emph{Relevant} or \emph{Irrelevant}, depending on their similarity with $q_o$. \emph{PerfectMatch} and \emph{Relevant} are considered as relevant. A question is composed of a subject, a body and a unique identifier. 

\subsection{Support Vector machines}

A top-performing model in the SemEval challenge is built with SVMs, which learn a classification function,  $f:Q \times Q \rightarrow \{0, 1\}$, on the relevant vs.~irrelevant questions belonging to the question set, $Q$. The classifier score is used to rerank a set of candidate questions  $q_{i}$ provided in the dataset with respect to an original question $q_{o}$. Three main representations were proposed: 
\Ni vectors of similarity features derived between two questions; 
\Nii a TK function applied to the syntactic structure of question pairs; or \Niii a combination of both. \vspace{.3em}

\noindent \textbf{Feature Vectors (FV)} are built for question pairs, $(q_1, q_2)$, using a set of \textit{text similarity} features that capture the relations between two questions. More specifically, we compute 20 similarities $sim( q_1, q_2)$ using word $n$-grams ($n=[1,\ldots,4]$), after stopword removal, greedy string tiling~\cite{Wise:1996}, longest common subsequences~\cite{Allison:1986}, Jaccard coefficient~\cite{Jaccard:1901}, word containment~\cite{Lyon:2001}, and cosine similarity.\vspace{.3em}

\noindent \textbf{Tree Kernels (TKs)} measure the similarity between the syntactic structures of two questions. Following \cite{S16-1172}, we build two macro-trees, one for each question in the pair, containing the syntactic trees of the sentences composing a question. In addition, we link two macro-trees by connecting the phrases, e.g., NP, VP, PP, etc., when there is a lexical match between the phrases of two questions. We apply the following kernel to two pairs of question trees: $K({\langle q_{1}, q_{2} \rangle}, {\langle q'_{1}, q'_{2} \rangle} ) = 
 TK(t(q_{1},q_{2}),t(q'_{1},q'_{2}))$+$TK(t(q_{2},q_{1}),t(q'_{2},q'_{1}))$, where $t(x,y)$ extracts the syntactic tree from the text $x$, enriching it with relational tags (REL) derived by matching the lexical between $x$ and $y$.

\vspace{-.3em}
\section{Injecting Structures in NNs}
\vspace{-.3em}
\label{sec:network}
We inject TK knowledge in two well-known and state-of-the-art networks for question similarity, enriching them with relational information.

\vspace{-.3em}
\subsection{NNs for question similarity}
\vspace{-.3em}

We implemented the Convolutional NN (CNN) model proposed by \cite{severyn2016modeling}. This learns $f$, using two separate sentence encoders $f_{q_1}:Q \rightarrow \mathbb{R}^n$ and $f_{q_2}:Q \rightarrow \mathbb{R}^n$, which map each question into a fixed size dense vector of  dimension $n$. The resulting vectors are concatenated and passed to a Multi Layer Perceptron that performs the final classification. Each question is encoded into a fixed size vector using an embedding layer, a convolution operation and a global max pooling function.
The embedding layer transforms the input question, i.e., a sequence of token, $X_{q} = [x_{q_1}, ...,x_{q_i} ,..., x_{q_n}] $, into a sentence matrix, $S_{q} \in R^{m \times n}$, by concatenating the word embeddings $w_i$ corresponding 
to the tokens $x_{q_i}$ in the input sentence.


Additionally, we implemented a Bidirectional (BiLSTM), using the standard LSTM by \citet{hochreiter1997long}. 
An LSTM iterates over the sentence one word at the time by creating a new word representation $h_i$ by composing the representation of the previews word and the current word vector $h_i = LSTM(w_i, h_{i-1})$. A BiLSTM iterates over the sentence in both directions and the final representation is a concatenation of the hidden representations, $h_N$, obtained after processing the whole sentence. We apply two sentence models (with different weights), one for each question, then we concatenate the two fixed-size representations and fed them to a Multi-Layer Perceptron.

\vspace{-.3em}
\subsection{Relational Information} 
\vspace{-.3em}
\citet{severyn2016modeling} showed that relational information encoded in terms of overlapping words between two pairs of text can highly improve accuracy. Thus, for both networks above, we mark each word with a binary feature indicating if a word from a question appears in the other pair question. This feature is encoded with a fixed size vector (in the same way it is done for words).

\vspace{-.3em}
\subsection{Learning NNs with structure}
\vspace{-.3em}
\label{sec:train}

%
To inject structured information in the network, we use a weak supervision technique: \Ni an SVM with TK is trained on the GS data; \Nii this model classifies an additional unlabelled dataset, creating automatic data; and \Niii a neural network is trained on the latter data. 

The pre-trained network can be fine-tuned on the GS data, using a smaller learning rate $\gamma$. This prevents catastrophic forgetting~\cite{goodfellow2014empirical}, which may occur with a larger learning rate.

\vspace{-.3em}
\section{Experiments}
\vspace{-.3em}

We experiment with two datasets comparing models trained on gold and automatic data and their combination, before and after fine tuning.
\vspace{-.3em}
\subsection{Data}
\vspace{-.3em}
\textbf{Quora dataset} contains $384,358$ pairs in the training set and $10,000$ pairs both in the dev. and test sets. The latter two contain the same number of positive and negative examples.

\noindent \textbf{QL dataset} contains $3,869$ question pairs divided in $2,669$, $500$ and  $700$ pairs in the training, dev.~and test sets. We created $93$k\footnote{Note that we will release the 400k automatically labelled pairs from Quora as well as the new $93$k pairs of QL along with their automatic labels for research purposes.} unlabelled pairs from the QL dump, retrieving 10 candidates with Lucene for $9,300$ query questions.


\begin{table*}[t]
\centering
\scalebox{0.90}{
\begin{tabular}{lcccc}
\toprule
Model       		& Automatic data & GS data & DEV    & TEST   \\ \midrule
FV-10k      	& --       & 10k   & 0.7046 & 0.7023 \\
TK-10k      	& --         & 10k   & 0.7405 & 0.7337 \\
CNN-10k      	& --         & 10k   & 0.7646 & 0.7569 \\
LSTM-10k      	& --         & 10k   & 0.7521 & 0.7450 \\
\midrule
CNN(CNN-10k)  & 50k      & --      & 0.7666 & 0.7619 \\
CNN(CNN-10k)* & 50k      & 10k   & 0.7601 & 0.7598 \\
CNN(FV-10k)  & 50k      &  --     & 0.6960 & 0.6931 \\
CNN(FV-10k)* & 50k      & 10k   & 0.7681 & 0.7565 \\
\midrule
CNN(TK-10k)  & 50k      &   --    & 0.7446 & 0.7370 \\
CNN(TK-10k)* & 50k      & 10k   & 0.7748 & 0.7652 \\
LSTM(TK-10k)  & 50k      &   --    & 0.7478 & 0.7371 \\
LSTM(TK-10k)* & 50k      & 10k   & 0.7706 & 0.7505 \\\midrule
TK-5k       &      --    & 5k    & 0.6859 & 0.6774 \\
CNN-5k       &    --      & 5k    & 0.7532 & 0.7450  \\
CNN(TK-5k)   & 50k      &   --    & 0.7239 & 0.7208 \\
CNN(TK-5k)*   & 50k      & 5k    & 0.7574 & 0.7493 \\ \midrule
CNN(TK-10k)  & 375k     &   --    & 0.7524 & 0.7471 \\
\textbf{CNN(TK-10k)*} & 375k     & 10k   & \textbf{0.7796} & \textbf{0.7728} \\
\bottomrule
\bottomrule
Voting(TK+CNN) & -- & 10k & 0.7838 & 0.7792 \\
\end{tabular}}
\vspace{-.3em}\caption{Accuracy on the Quora dataset.}
\vspace{-.5em}
\label{tab:quora}
\end{table*}

\vspace{-.5em}
\subsection{NN setup}
\vspace{-.3em}
We pre-initialize our word embeddings with skip-gram embeddings of dimensionality 50 jointly trained on the English Wikipedia dump~\cite{word2vec} and the jacana corpus\footnote{Embeddings are available in the repository: \url{https://github.com/aseveryn/deep-qa}}.
The input sentences are encoded with fixed-sized vectors using a CNN with the following parameters: a window of size 5, an  output of 100 dimensions, followed by a global max pooling. We use a single non-linear hidden layer, whose size is equal to the size of the sentence embeddings, i.e., 100. The word overlap embeddings is set to 5 dimensions.
The activation function for both convolution and hidden layers is ReLU. 
During training the model optimizes the binary cross-entropy loss. 
We used SGD with Adam update rule, setting the learning rate to $\gamma$ to $10^{-4}$ and $10^{-5}$ for the pre-training and fine tuning phases, respectively.

\vspace{-.3em}
\subsection{Results on Quora}
\vspace{-.1em}
Table~\ref{tab:quora} reports our different models, FV, TK, CNN and LSTM described in the previous section, where the suffix, -10k or -5k, indicates the amount of GS data used to train them, and the name in parenthesis indicates the model used for generating automatic data, e.g., CNN(TK-10k) means that a CNN has been pre-trained with the data labelled by a TK model trained on 10k GS data. The amount of automatic data for pre-training is in the second column, while the amount of GS data for training or fine tuning (indicated by $^*$) is in the third column. Finally, the results on the dev.~and test sets are in the fourth and fifth columns.

We note that: first, NNs trained on 10k of GS data obtain higher accuracy than FV and TK on both dev.~and test sets (see the first four lines); 

Second, CNNs pre-trained with the data generated by FV or in a self-training setting, i.e., CNN(CNN-10k), and also fine-tuned do not improve\footnote{The improvement of 0.5 is not statistically significant.} on the baseline model, i.e., CNN-10K, (see the second part of the table). 

Third, when CNNs and LSTMs are trained on the data labelled by the TK model, match the TK model accuracy (third part of the table). Most importantly, when they are fine-tuned on GS data, they obtain better results than the original models trained on the same amount of data, e.g., ~1\% accuracy over CNN-10k.  

Next, the fourth part of the table shows that the improvement given by our method is still present when training TK (and fine tuning the NNs) on less GS data, i.e., only 5k. 

\begin{figure}[t]
\vspace{-1.45em}
\hspace{-1em}
\includegraphics[scale=0.57]{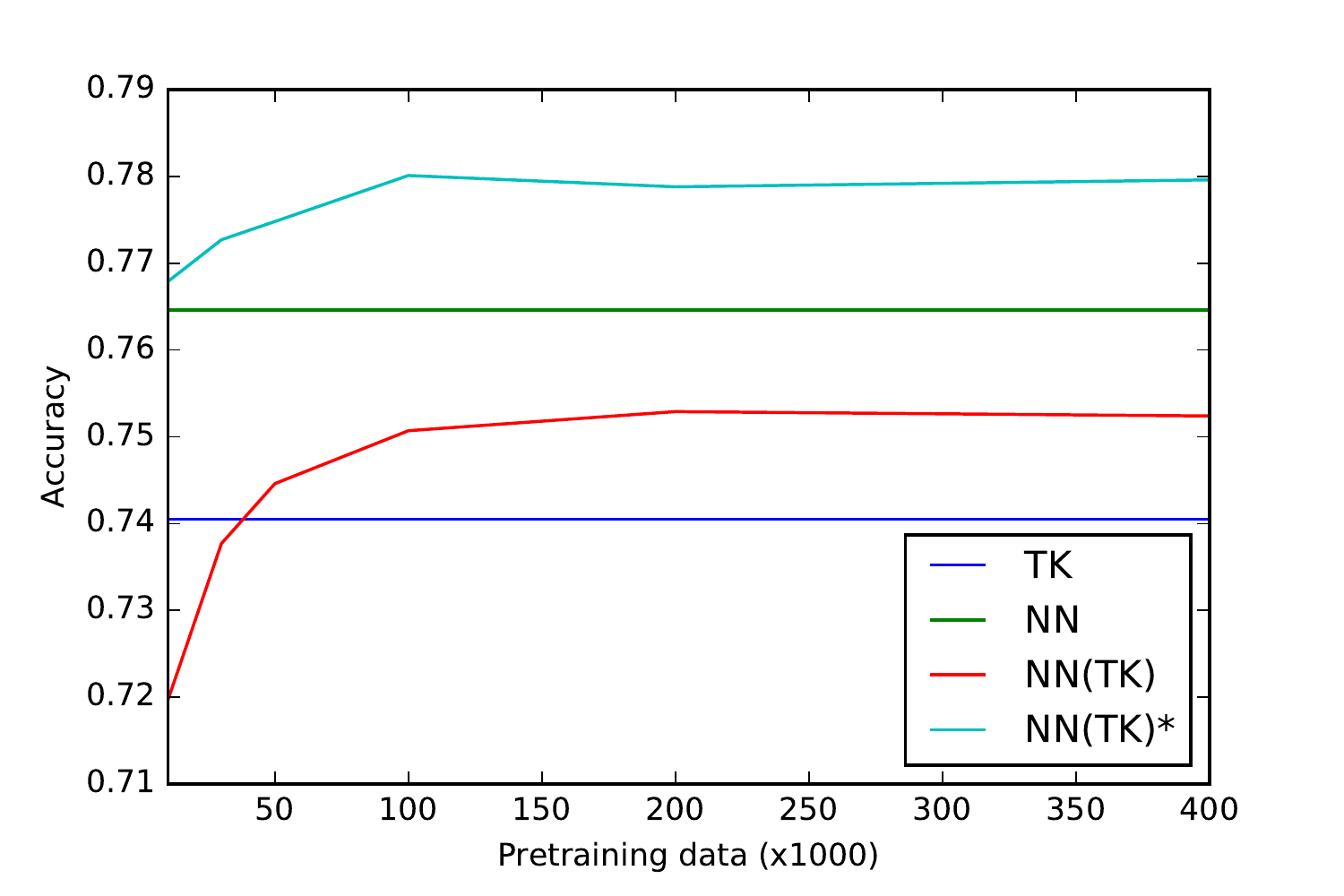}
\vspace{-2em}
\caption{Impact of the pre-training data.}
\label{fig:learning-curve}
\vspace{-.5em}
\end{figure}

Additionally, the fifth section of the table shows a high improvement by training NNs on all available Quora data  annotated by TK-10k (trained on just 10k). This suggests that NNs require more data to learn complex relational syntactic patterns expressed by TKs. However, the plot in Figure \ref{fig:learning-curve} shows that the improvement reaches a plateau around 100k examples. 

Finally, in the last row of the table, we report the result of a voting approach using a combination of the normalized scores of TK-10k and CNN-10k. The accuracy is almost the same than CNN(TK-10k)*. This shows that NNs completely learn the combination of a TK model, mainly exploiting syntax, and a CNN, only using lexical information. Note that the voting model is heavy to deploy as it uses syntactic parsing and the kernel algorithm, which has a time complexity quadratic in the number of support vectors.

\subsection{Results on Qatar Living}

Table~\ref{tab:ql} reports the results when applying our technique to a smaller and different dataset such as QL.
Here, CNNs have lower performance than TK models as 2,669 pairs are not enough to train their parameters, and the text is also noisy, i.e., there are a lot of spelling errors. Despite this problem, the results show that CNNs can approximate the TK models well, when using a large set of automatic data.
For example, the CNN trained on 93k automatically annotated examples and then fine tuned exhibits 0.4\% accuracy improvement on the dev.~set and almost 3\% on the test set over TK models.
On the other hand, using too much automatically labeled data may hurt the performance on the test set. This may be due to the fact the quality of information contained in the gold labeled data deteriorates. In other words, using the right amount of weekly-supervision is an important hyper-parameter that needs to be carefully chosen.

\section{Related Work}

Determining question similarity is one of the main challenges in building systems that answer real user questions \cite{agichtein2015overview,Agichtein2016OverviewOT} in community QA, thus different approaches have been proposed.
\citet{jeon2005finding} used a language model based on word translation table to compute the probability of generating a query question, given a target/related question.  
\citet{zhou2011phrase} showed the effectiveness of phrase-based translation models on Yahoo!~Answers. 
\citet{cao2009use,duan2008searching} proposed a similarity between two questions based on a language model that exploits the category structure of Yahoo! Answers. 
%
\citet{wang2009syntactic} proposed a model to find semantically related questions by computing similarity between syntactic trees representing questions. 
\citet{ji2012question} and \citet{zhang2014question} used latent semantic topics that generate question/answer pairs. 

Regarding the use of automatically labelled data,
\citet{blum1998combining} applied semi-supervised approaches, such as self-training and co-training to non-neural models. The main point of our paper is the use standard weakly-supervised methods to inject syntactic information in NNs.

\citet{P16-1228} tried to combine symbolic representations with NNs by transferring structured information of logic rules into the weights of NNs. Our work is rather different as we inject syntactic, and not logic, information in NNs.

The work most similar to our is the one by \citet{croce2017deep}, who use Nystrom methods to compact the TK representation in embedding vectors and use the latter to train a feed forward NNs. In contrast, we present a  simpler approach, where NNs learn syntactic properties directly from data.

To our knowledge, ours is the first work trying to use NNs to learn structural information from data labelled by TK-based models. Finally, no systems of the SemEval challenges used NNs trained on syntactic information.


\begin{table}[t]
\centering
\scalebox{0.9}{
\begin{tabular}{lccHcH}
\toprule
Model   		& Automatic Data 	& Dev 	& Dev (MAP)  & Test  & Test (MAP)  \\ \midrule
CNN      		&             		& 0.7000 	& 0.6598& 0.7514 &0.7208\\
TK      		&            		& 0.7340 	& \textbf{0.6988}&0.7686 &0.7424\\
\midrule
CNN(TK)  		& 50k        	& 0.5580 	& 0.6578&0.5428 &0.7370\\
CNN(TK)* 	& 50k        	& 0.7160	 & 0.6794&\textbf{0.7814} &0.7312\\ 
CNN(TK)  		& 93k      	& 0.7000 	& 0.6782 &0.6957& \textbf{0.7430} \\
CNN(TK)* 	& 93k      	& \textbf{0.7380} &0.6782 &0.7614 &0.7320 \\
\bottomrule
\end{tabular}
}
\vspace{.3em}
\caption{Accuracy on QL using all available GS data.}
\label{tab:ql}
\vspace{-.5em}
\end{table}

\section{Conclusion}

In this work, we have trained TK-based models, which make use of structural information, on relatively small data and applied them to new data to produce a much larger automatically labeled dataset. Our experiments show that NNs trained on the automatic data improve their accuracy. We may speculate that NNs learn relational structural  information as \Ni TK models mainly use syntactic structures to label data and \Nii other advanced models based on similarity feature vectors do not produce any improvement. Indeed, the latter only exploit lexical similarity measures, which are typically also generated by NNs. However, even if our conjecture were wrong, the bottom line would be that, thanks to our approach, we can have NN models comparable to TK-based approaches, by also avoiding to use syntactic parsing and expensive TK processing at deployment time.




\bibliography{naaclhlt2018}
\bibliographystyle{acl_natbib}

\end{document}